\documentclass{article}

\usepackage{PRIMEarxiv}
\usepackage{makecell}
\usepackage{tabularx}
\usepackage[utf8]{inputenc} % allow utf-8 input
\usepackage[T1]{fontenc}    % use 8-bit T1 fonts
\usepackage[hidelinks,urlcolor=blue]{hyperref}
\usepackage{url}            % simple URL typesetting
\usepackage{booktabs}       % professional-quality tables
\usepackage{amsfonts}       % blackboard math symbols
\usepackage{nicefrac}       % compact symbols for 1/2, etc.
\usepackage{microtype}      % microtypography
\usepackage{lipsum}
\usepackage{fancyhdr}       % header
\usepackage{graphicx}       % graphics
\graphicspath{{}}     % organize your images and other figures under media/ folder

\title{The Model Openness Framework: Promoting Completeness and Openness for Reproducibility, Transparency, and Usability in Artificial Intelligence}

\author{
  Matt White\thanks{Corresponding author: Matt White, matt.white@linuxfoundation.org} \\
  Linux Foundation \\
  San Francisco, CA, USA\\
   \And
  Ibrahim Haddad \\
  Linux Foundation \\
  San Francisco, CA, USA\\
 \AND
  Cailean Osborne \\
  University of Oxford \\
  Oxford, UK\\
   \And
  Xiao-Yang Yanglet Liu\\
  Columbia University \\ 
  New York, USA\\
 \AND
  Ahmed Abdelmonsef \\
  Generative AI Commons \\
  Cairo, Egypt\\
   \And
  Sachin Mathew Varghese \\
  Generative AI Commons \\
  Jacksonville, FL, USA\\
   \And
  Arnaud Le Hors \\
  IBM \& Generative AI Commons \\
  San Jose, CA, USA\\
}

\begin{document}
\maketitle
%TC:ignore
\vspace{-1em}

\begin{abstract}
Generative artificial intelligence (AI) offers numerous opportunities for research and innovation, but its commercialization has raised concerns about the transparency and safety of frontier AI models. Most models lack the necessary components for full understanding, auditing, and reproducibility, and some model producers use restrictive licenses whilst claiming that their models are ``open source''. To address these concerns, we introduce the Model Openness Framework (MOF), a three-tiered ranked classification system that rates machine learning models based on their completeness and openness, following open science principles. For each MOF class, we specify code, data, and documentation components of the model development lifecycle that must be released and under which open licenses. In addition, the Model Openness Tool (MOT) provides a user-friendly reference implementation to evaluate the openness and completeness of models against the MOF classification system. Together, the MOF and MOT provide timely practical guidance for (i) model producers to enhance the openness and completeness of their publicly-released models, and (ii) model consumers to identify open models and their constituent components that can be permissively used, studied, modified, and redistributed. Through the MOF, we seek to establish completeness and openness as core tenets of responsible AI research and development, and to promote best practices in the burgeoning open AI ecosystem.
\end{abstract}


%TC:endignore
\section{Introduction}
Artificial intelligence (AI) has seen remarkable advances in recent years \cite{lecun2015deep}, driven by the growth in computational capabilities \cite{dean2018new,AIchips2022}, available training data \cite{mayer2013big,deng2009imagenet}, and improved deep learning algorithms \cite{goodfellow2016deep,ho2020denoising,vaswani2017attention}. However, as AI systems have become more advanced, concerns have also grown regarding their transparency, reproducibility, and safety \cite{buolamwini2018gender,weidinger2021ethical,bender2021dangers}. Most state-of-the-art (SOTA) models are black boxes, making it hard to explain their internal logic or to ensure fairness \cite{castelvecchi2016can,bommasani2021opportunities}. While the number of publicly available models has been growing, many of these models are falsely being promoted as ``open-source'', a practice that has been characterized as  ``openwashing'' \cite{widder2023open,nolan2023llama,liesenfeld2023opening,liesenfeld2024rethinking}. The lack of transparency and reproducibility in AI models hinders scientific progress and erodes trust in AI research and development (R\&D) \cite{hutson2018ai,pineau2021improving}. Without a standardized framework to evaluate and promote openness, it becomes challenging to verify claims, build upon existing work, and ensure responsible development. 

To address these concerns, we introduce the Model Openness Framework (MOF) for evaluating and classifying the completeness and openness of machine learning (ML) models across their development process. Model producers must go beyond releasing models and trained weights; they should include all artifacts involved in the model development lifecycle. The MOF contributes to broader efforts that seek to promote transparency, reproducibility, and responsibility in AI R\&D, including reproducibility checklists \cite{pineau2021improving}, ethical AI guidelines \cite{jobin2019global}, model and data cards \cite{mitchell_modelcards_2019,gebru2021datasheets}. By adopting the MOF, the AI community can create a more open, accountable, and trustworthy ecosystem.

For the sake of simplicity in nomenclature, this paper refers to any person or entity that develops and trains a first-generation model as a ``model producer'' or simply a ``producer''. This encompasses AI researchers, developers, AI hobbyists, or anyone who trains a model in some form or fashion, including fine-tuning and alignment, as long as they are the originator of the (foundation) model. Similarly, any person or entity that adopts, consumes, alters, or uses a model and corresponding artifacts for any purpose including modifying weights through fine-tuning is referred to as a ``model consumer'' or simply a ``consumer''. This includes end users, researchers, developers, or anyone that uses an ML model and is not its producer. We also use the terms ``ML'' and ``ML model'' to broadly describe any model, whether classical machine learning or deep learning and both generative and discriminative. 

The paper has the following structure. It begins with a discussion of related work and how the MOF builds on prior approaches to evaluating the openness of models (Section~\ref{sec:relatedwork}). Next, it reviews the concepts of openness and completeness in science and technology (Section~\ref{sec:completenessopenness}). Then, it introduces the three classes of the MOF classification system (Section~\ref{sec:MOFClasses}), as well as the 17 components (Section~\ref{sec:mofcomponents}) and acceptable licenses per component (Section~\ref{sec:moflicenses}). Then, it discusses how to adopt the framework in practice (Section~\ref{sec:mofimplementation}), as well as the benefits (Section~\ref{sec:mofbenefits}) and limitations (Section~\ref{sec:moflimitations}) of the MOF. Finally, it concludes with a summary of the key contributions for both model producers and consumers (Section~\ref{sec:conclusion}).

\section{Related Work} \label{sec:relatedwork}

\subsection{Opening the Black Box: Benefits and Risks of Openness in AI}
While AI has seen remarkable advances in recent years \cite{lecun2015deep}, most SOTA foundation models are black boxes, making it hard to audit or explain their logic or behavior \cite{castelvecchi2016can,bommasani2021opportunities}. Large language model services like OpenAI's GPT-4 hide opaque models behind cloud-based APIs, providing no insight into the inner workings \cite{openai2024gpt4}.  To address these concerns, there has been a growing movement towards the openness of models with companies, research organizations, and individuals sharing models on platforms like Hugging Face Hub, GitHub, and Kaggle \cite{phang2022eleutherai,castano2024lessons,osborne2024ai}. Furthermore, grassroots initiatives have emerged as the early leaders in the open development of open foundation models \cite{ding2023towards, akiki2022bigscience}, such as GPT-Neo by EleutherAI \cite{black2022gpt}, BLOOM by BigScience \cite{workshop2022bloom}, and SantaCoder by BigCode \cite{allal2023santacoder}. This shift towards the open development of AI models is increasingly viewed as a credible alternative to closed-source development \cite{patel2023google}.

There has been much debate about the benefits and risks of releasing models \cite{solaiman2019release,seger2023open,kapoor2024societal,solaiman2023gradient,law2023open,eiras2024risks}. On the one hand, the accessibility and transparency of open models can concurrently deliver advantages over closed source models, including security and performance advantages through distributed development and auditing \cite{WladawskyBerger2023,raji2019actionable}, adaptability and customization for diverse domains and languages \cite{kapoor2024societal,pipatanakul2023typhoon}, as well as advances in science \cite{yang2023shadow,kirchenbauer2023watermark,han2023medalpaca}. On the other hand, the openness of models introduces a number of risks, such as the generation of disinformation \cite{musser2023cost,menczer2023addressing}, illegal content \cite{thiel2023generative}, as well as security vulnerabilities \cite{tsamados2023cybersecurity}. Open foundational models are understood to have five distinctive properties that present \textit{both} benefits and risks: broader access, greater customizability, local adaptation and inference ability, the inability to rescind model access, and the inability to monitor or moderate model usage \cite{kapoor2024societal}. A systematic review of benefits and risks in the short, medium, and long terms concludes that the benefits outweigh the costs and accordingly encourages the open sourcing of models as well as training and evaluation data \cite{eiras2024risks}. Striking a balance between the benefits and associated risks of open models remains a critical challenge in the AI R\&D landscape.

\subsection{Lack of Openness in ``Open Source'' AI}

ML models whose weights are made publicly-available for download and downstream use are being falsely promoted as ``open-source'' \cite{nolan2023llama,maffulli2023meta,liesenfeld2023opening}. Such models may more accurately be described as ``open-weight models'' \cite{liesenfeld2024rethinking}.  While there is a fast-growing number of open models and open datasets shared on online platforms, a concerning number of models and datasets are shared either without licenses---for example, 64.67\% of models and 72.13\% of datasets on Hugging Face Hub are unlicensed \cite{osborne2024ai}---or with restrictive licenses that do not meet the standards required of open licenses \cite{nolan2023llama,maffulli2023meta}. In most cases, pretraining data or human feedback data collected from usage of models in, for example, chatbots are not released \cite{donyehiya2024}. Some model producers even add conditions that stipulate that their model outputs cannot be used to train subsequent models or add trigger conditions that would require a model consumer to negotiate a new license when some condition is met. %Projects of this nature limit innovation and create risks for organizations seeking to use them for various purposes, forcing them to navigate restrictions that make them unsuitable for broad adoption. 
In addition, fine-tuned models based on foundation models with restrictive licenses are being released with open-source licenses, such as Apache 2.0, even though altering the original license is not legally permitted. This creates confusion in the ecosystem and can have legal consequences for those altering the license and those using the model. 

Many open (foundation) models are released with technical reports and model cards that provide limited information on the source and treatment of their training data, fine-tuning, or alignment methods \cite{openai2024gpt4,jiang2024mixtral}, and evaluation results often cannot be reproduced independently due to the lack of their disclosure \cite{mcintosh2024inadequacies}. Furthermore, few disclosures are made about guardrails and if prompts and outputs are altered, filtered, or replaced \cite{Mills2023,StokelWalker2023}.  Overall, the lack of openness leaves downstream model consumers to rely on limited claims reported by the model producers.

The misrepresentation of models as ``open source'' is in part due to confusion about the appropriate use of open-source licenses. Many developers do not realize that open-source licenses were designed to cover conventional software code and are not appropriate for the intricacies of ML models \cite{OSI2007,OSI_AI_2023}. As we discuss in Section~\ref{sec:moflicenses}, open-source licenses cover the model architecture, which is defined in software code, but not the corresponding model parameters. By contrast, model parameters are data and are more aptly governed by open-data licenses than by open-source licenses \cite{OpenKnowledgeFoundation2018}. Meanwhile the misrepresentation of models as ``open source'' by companies has also been characterized as ``openwashing'' \cite{nolan2023llama,liesenfeld2023opening,liesenfeld2024rethinking}, where ``open'' has been used imprecisely and loosely to describe ``systems that offer minimal transparency or reusability…alongside those that offer maximal transparency, reusability, and extensibility'' \cite{widder2023open}. This problem motivates our creation of a ranking system to promote openness and completeness.
%Furthermore, a review of large language models found that, ``[W]hile there is a fast-growing list of projects billing themselves as 'open source', many inherit undocumented data of dubious legality, few share the all-important instruction-tuning..., and careful scientific documentation is exceedingly rare''  \cite{liesenfeld2023opening}. 

Another challenge is that most models have fallen short in their completeness (i.e. the full availability of components from the model development lifecycle, see list in Section~\ref{sec:mofcomponents}), only releasing model architectures and final trained parameters. To achieve full transparency, reproducibility, and extensibility, we argue that model producers must go beyond just releasing their model and the trained weights and biases, which is currently the norm. Instead, they should include all artifacts of their work, including datasets for training, validation, testing, and benchmarking, as well as detailed documentation, such as research papers, model cards, data cards, and any usage documentation. Completeness also requires all code used to parse and process data, the code used for training and inference, and any code used in benchmark tests, along with any libraries or other code artifacts that were a part of the model development lifecycle.

\subsection{MOF: A Novel Approach to Evaluating Model Openness \& Completeness}
There is not yet a formally agreed-upon definition of ``open source AI'' \cite{OSI_AI_2023}. Broadly, open AI refers to the concept of transparency and accessibility in AI R\&D. It entails the sharing of key artifacts associated with the development of models, including data, code, models, and publications, under both open and restrictive licenses, which allow access, inspection, modification, or distribution of models. As mentioned above, open AI also entails grassroots initiatives that have used open collaboration approaches to develop open-weight models  \cite{ding2023towards, akiki2022bigscience}. The sharing of models grants the community the freedoms to transparently review capabilities and limitations, identify issues, reuse or extend functionality, and participate in collective advancement. This is enabled through open licenses applied judiciously to key model components, including datasets, model architectures, and trained parameters, which facilitates attribution, safeguards model consumers, and maintains community norms while removing barriers to adoption \cite{gonzalez2005open,lin2006open}.  

The combination of open source, open data, open access, and open science is a powerful and effective way of solving the most pressing issues in AI R\&D, including access, explainability, transparency, reproducibility, and safety. The goals of open AI are to accelerate progress through open collaboration, establish trust by allowing system inspection, enable diverse perspectives, and align AI advancement with social benefits \cite{phang2022eleutherai}. Due to the nascent nature of the open AI movement, new standards are being developed to address shortcomings, including the draft Open Source AI Definition \cite{OSI_AI_2023}; tools for auditing model explainability, fairness, and robustness \cite{birhane2024ai,guidotti2018survey,raji2020closing,richardson2021framework,arya2019one,mokander2023auditing}; frameworks to evaluate model openness, such as the AAAI Reproducibility Checklist \cite{nosek_openresearch_2015} and the NeurIPS 2019 ML Reproducibility Checklist \cite{pineau2021improving}; the establishment of ethics review boards in AI research labs \cite{schuett2024design}; as well as work by government agencies, including NIST and NTIA in the USA \cite{NTIA2023} and the AI Safety Institute in the UK \cite{aisi-inspect}.

However, prior approaches do not evaluate both the completeness and openness of models. The MOF reinforces existing approaches by objectively evaluating and classifying models based on which components of the development lifecycle are released under open-source licenses. It codifies openness across model development pipelines with informative guidelines, a classification system, and a method for assigning badges to qualified models. Models with licenses that do not impose downstream restrictions are considered open, while restrictive ones are source-available. This differs from the gradient approach to model openness \cite{solaiman2023gradient}, which classifies BLOOM by BigScience \cite{workshop2022bloom} and GPT-J by EleutherAI \cite{mesh-transformer-jax} as open. We would classify GPT-J as open because it was released under the OSI-approved Apache 2.0 license, while BLOOM is source-available due to its restrictive, non-OSI-approved OpenRAIL license \cite{ferrandis2022openrail}. Overall, the MOF encourages model producers to strive for complete transparency and usability without restrictions.

\section{Understanding the Concepts and Culture of Openness and Completeness} \label{sec:completenessopenness}
Before presenting the details of the MOF, we review the concepts of openness and completeness in science and technology. These core tenets form the basis of open science, open source, open data, and open access, which enable transparency, reproducibility, and collaboration in research, and are part of the wider open knowledge movement that believes all knowledge should be shared freely \cite{White2023A}. This section provides an overview of each domain and how they connect to the framework's goals. Understanding the motivations behind openness clarifies why it is vital to extend these concepts to AI R\&D: it facilitates the democratization of AI, which is essential for advancing AI research and innovation, as well as responsibility in AI R\&D, including transparency, accessibility, and inclusivity \cite{White2023b}.

\subsection{Openness}
Openness is the practice of freely sharing the methodology, progress, and products of R\&D with the public without restrictions on access, inspection, modification, or distribution \cite{peters2015virtues}. Instead of limiting transparency through proprietary terms, openness concerns the release of materials under permissive open licenses tailored to the type of content. This upholds scientific ideals around reproducibility, accountability, and cumulative innovation, while empowering research and developer communities to meaningfully review, discuss, reuse, and extend upon prior work \cite{vicente2018open}. As we elaborate in Section~\ref{sec:openlicenses}, the careful selection of appropriate open licenses facilitates attribution, protects downstream consumers, maintains community ethical norms, and facilitates adoption and impact \cite{colazo2009impact, lin2006open}. 

We also seek to differentiate between the terms ``open'' and ``complete'' in order to make it clear to model consumers exactly what model producers are providing and under what conditions when they say their model is open. Openness is not just about what is included, but importantly under what license each component is released. We believe opening the ``black box'' of AI will be crucial for continued advances and responsible use \cite{mittelstadt2019explaining}.  Although open-source licensing is imperative for the code components that are provided for the MOF, our approach to the MOF aligns with wider open science principles and the vision of open AI that requires more than open-source licenses for code components for models to be considered open. For example, non-code elements like datasets and research papers need an appropriate license that suits its format, such as open-data or open-content licenses, which are not currently OSI-approved licenses.

\subsection{Completeness}

Completeness is a core tenet of open science \cite{vicente2018open}. We define completeness as the availability of key artifacts produced during the full lifecycle of conducting research or the engineering of a technical product, enabling comprehensive transparency, inspection, evaluation, and reproducibility. In the context of ML, completeness entails releasing all the key components associated with developing an ML model rather than just selected artifacts. It entails sharing the full pipeline that produced a model in a usable form. Comprehensive releases empower unfettered scrutiny into model genetics: curation and treatment of training data, feature engineering, neural architectures, weight evolution, training configurations, model performance across diverse benchmarks, replication of model producer claims, and other byproducts of the model development lifecycle. The MOF encourages model producers to exhibit full completeness, providing all artifacts involved in the model development lifecycle when distributing models. It defines an ascending hierarchy of criteria for releasing key artifacts with the highest bar aligned with open science paradigms. Completeness combined with openness (open licensing) accelerates collective advancement of trustworthy and innovative AI.

We use the term ``completeness'' borrowed from open science to disambiguate from the multiple uses of the word ``openness'', which has unfortunately become a vague and confusing term \cite{widder2023open,liesenfeld2023opening}. Openness is often used to describe not only the licensing used for artifacts but also the availability of artifacts and even the thoroughness of those artifacts. The multiple uses of the term ``open'' continues to be used in a way that is misleading or does not reveal the specifics of its usage \cite{NTIA2023}. Packing the term ``openness'' with multiple definitions, uses, or dimensions does not clearly articulate what aspect of the model is open. For instance, a model producer may claim that their model is ``open'' but model consumers may not know if it is open because it employs open licenses, because it is made publicly available, because it provides additional components like datasets, or because the components released are thorough or usable. For this reason, we use the term ``completeness'' to measure the availability of components that are released with models (with the goal of full completeness) and the term ``openness'' to describe the usage of permissive licenses for components.

\subsection{Open Knowledge}
Open knowledge is an overarching philosophy and larger movement that encompasses all the preceding areas of openness, revolving around the free and public sharing of information and insights across various domains \cite{garcia2010open,molloy2011open}. This entails making knowledge resources accessible to everyone and contributing to a wider pool of shared understanding. Open knowledge practices also involve ensuring that the information is ethically curated and disseminated, upholding principles of integrity and respect for intellectual property. The Wikimedia Foundation, Open Knowledge Foundation, and Science Commons are leading organizations in the open knowledge community.

\subsection{Open Science}
Open science refers to the practice of making all stages of the scientific process transparent and accessible to others \cite{vicente2018open, chesbrough2015open}. This includes publishing research papers, data, source code, code notebooks, and any information or tools needed to replicate research. The goals of open science are to enable reproducibility, collaboration, and facilitate building on previous knowledge to advance scientific research \cite{vicente2018open}. Open science is critical for credible, ethical, and accessible scientific research that can be reviewed, validated, replicated, and built upon. Open science in AI is sometimes referred to as ``open science AI'' and is the gold standard for ensuring reproducibility and transparency.

Advances in AI R\&D are in part attributed to the sharing of preprints on platforms like arXiv, but much of the training data, model details, and code of SOTA AI systems remain proprietary. The opaque nature of many AI systems limits reproducibility, hinders research, and increases concerns around bias and safety. Transitioning to open datasets, architectures, weights, and code promises to facilitate AI research, innovation, and adoption across the private and public sectors. Overall, openness has repeatedly shown immense power to advance progress, equity, and opportunity across endeavors. The MOF aims to promote the spirit and methodology of open science in the AI R\&D community.

\subsection{Open Access}
Open access is the process of making research outputs like publications freely available to read without subscriptions or paywalls, enabling broad dissemination of knowledge.  \cite{suber2012open, tennant2016academic}. There are various open-access platforms like Cornell University’s arXiv, which make publications, often distributed under an open license, freely available for review. Furthermore, the adoption of open access policies, mandates, and licenses by journals and conferences have contributed to greater access to research. Before open access, research publications were mostly locked behind expensive journal subscriptions and paywalls, which limited the discoverability and use of knowledge. The open access movement has made more research freely available to all. Open access speeds the dissemination of discoveries to scientists and the public, and it facilitates reproducibility and meta research. As a result, entry barriers to accessing research have greatly reduced and public access to AI research papers has helped advance the field, including many of the developments and enhancements to the transformer architecture that powers the latest highly-capable LLMs.

\subsection{Open Collaboration and Open Community}
Open collaboration encourages cooperative efforts across institutions, disciplines, and borders, involving more inclusive and diverse participation in the development of science and technology \cite{enkel2009open,chesbrough2015open,chesbrough2003open}. Open community goes beyond open collaboration, and it concerns the creation and sustainability of a shared community with neutral governance, where projects can be worked on collaboratively in an equitable environment that embraces principles of openness. The LF AI \& Data and Generative AI Commons are examples of open communities \cite{Kerner2023}. 

\subsection{Open Source Software}
Open source software (OSS) involves publishing software code under licenses that grant users independence and control over the technology by allowing inspection, modification, and redistribution of the code without restrictions \cite{OSI2007}. The OSS movement has transformed software development over the past few decades: while early closed and proprietary systems limited access, locked in users, and stagnated innovation \cite{bretthauer2001open}; nowadays OSS is estimated to be used in 96\% of global code bases \cite{synopsys_open_2023} and to constitute up to 90\% of software stacks \cite{open_source_security_foundation_oss_2022}. It is increasingly being recognized as digital infrastructure \cite{eghbal_roads_2016,scott_avoiding_2023}. OSI-approved licenses like Apache 2.0 and MIT have been key to enabling worldwide collaborative development, freedom of choice, and accelerated progress \cite{gonzalez2005open}. 

OSS has emerged as an indispensable component of AI R\&D \cite{langenkampyue2022,osborne2024publicprivate} and open science at large \cite{willinsky2005unacknowledged,von2007open}. OSS presents a myriad of benefits for individuals \cite{von2012carrots,lakhani2005hackers} and enterprises \cite{bonaccorsi2006comparing,li2024systematic}. It encourages the sharing of code and software development methodologies \cite{lerner2002some}, providing a basis for building upon existing work and contributing to the advancement and democratization of science \cite{sonnenburgml2007,schweik2006free,osborne2024AIdemo}; it provides learning, skill development, and career development opportunities \cite{ding2023towards, alexander2002working}; it reduces software development and testing costs \cite{Chesbrough_2023,lindman2009beyond, birkinbine2020incorporating}; and it facilitates the development of open standards \cite{lerner2002some,fink2003business}, among others. The benefits of OSS do not come without risks, especially at the hand of bad actors and in light of vulnerabilities \cite{hiesgen2022race,payne2002security}. Yet a wide user base and the sharing of knowledge enables OSS projects to quickly identify vulnerabilities and fix issues. A good example is the identification of the recent XZ attack (CVE-2024-3094) and public reporting by the open source community \cite{Bals_2024,Naughton_2024}.

\subsection{Source Available}
Source available should not be confused with open source. Source available originated from conventional software development, where a developer provides access to the source code, but the licenses are not open-source. This means they include restrictions that consumers must fully understand before agreeing to use it. Some have referred to these projects as open access, but this is a misnomer since open access applies to documentation without paywalls. Most open-washed projects are examples of source available due to their restrictive licensing \cite{nolan2023llama,maffulli2023meta}.

\subsection{Open Data}

Open data refers to the public release of datasets, databases, and other structured data used for research, enabling access and reuse \cite{murray2008open,kitchin2014data}. This practice upholds scientific reproducibility, allows reanalysis, and spurs innovation \cite{heltweg2023}. Standard policies and formats are often employed to ensure quality and usable data sharing. Open content, on the other hand, refers to the sharing of creative materials and unstructured data. Both open-data and open-content licenses exist, with open-data licenses often applicable to both data and content. Open data emphasizes the standardization of datasets, addressing transparency and requiring comprehensive descriptions of data collection methods and assessments for intrinsic bias. Furthermore, accessibility is a cornerstone of open data, with datasets expected to be readily available without personal requests or paywalls, promoting transparency and enabling scrutiny. 

Historically, many research fields had cultures of data secrecy that impaired reproducibility and knowledge building. Openly sharing data enables reanalysis, reproducibility, and new applications \cite{uhlir2007open,murray2008open}. For instance, government open data initiatives provide transparency of government operations and promote innovation in and for the public sector \cite{janssen2012benefits, jaakkola2014open}; opening clinical trial data facilitate pharmaceutical research \cite{krlevza2016impact} and open genomic databases enablee bioinformatics breakthroughs \cite{knoppers2003human}; and open climate data \cite{reichman2011challenges,de2020open} have fuelled research and innovation to combate climate change. While better standards and tooling around open data publishing are still needed, the value of open data is clear. In the context of AI R\&D, the Datasets and Benchmarks track at NeurIPS underscores the paramount importance of openly releasing machine learning datasets \cite{gebru2021datasheets}. 

\subsection{Open Licenses} \label{sec:openlicenses}
Open licenses are legal mechanisms that allow content and artifacts to be freely accessed, used, modified, and shared under permissive terms. They are essential for operationalizing open science, open data, and open-source ideals \cite{gonzalez2005open}. Different licenses have emerged for addressing rights, responsibilities, and permissible usage for data, publications, code, and other research outputs. Open licenses solve key problems with closed, restricted systems, including:
\begin{itemize}
    \item Enabling free access without paywalls or subscriptions
    \item Allowing reproduction, analysis, and extension of work
    \item Disseminating contributions back to the community
    \item Progressing cumulatively by building on prior ideas
    \item Fostering collaboration across organizational and geographic boundaries
    \item Promoting transparency and accountability
    \item Mitigating anti-competitive behavior or rent-seeking
\end{itemize}

For research papers and scholarly works, Creative Commons (CC) licenses are widely adopted, which allow free distribution and reuse with conditions, such as requiring attribution and allowing commercial use and derivative works. Common choices for open licenses are CC-BY (attribute) and CC-BY-SA (share alike). Using permissive CC licenses for papers, technical reports, and documentation provides rights to reproduce, expand, and translate the works \cite{lin2006open}.

For software code, many open-source licenses have been developed. The Open Source Definition and the list of approved open-source licenses is maintained by the OSI \cite{OSI2007}. Using OSI-approved open-source licenses encourages community review and contributions to code, promoting quality and shared progress \cite{colazo2009impact}. Prominent examples include the MIT, Apache 2.0, and the 3-Clause BSD license, which allow inspection, modification, and redistribution of code while requiring preservation of copyright and license terms. Alternative licenses, such as the Llama 2 license, OpenRAIL, and AI2 ImpACT licenses, are not considered open-source licenses due to their restrictions on usage \cite{maffulli2023meta}. 

For datasets, typical licenses are Creative Commons licenses, particularly Creative Commons Zero (CC0), CC BY (attribution) and CC BY-SA (Attribution-ShareAlike), as well as Linux Foundation’s Community Data License Agreement (CDLA-Permissive) and the Open Data Commons licenses like Public Domain Dedication and License (PDDL) and the Open Data Commons Attribution License License (ODC-By). They provide terms for sharing data openly while addressing concerns, such as attribution, permissive usage, and liability  \cite{lin2006open}. 

\section{Model Openness Framework Classes} \label{sec:MOFClasses}
\subsection{MOF Structure}
The MOF proposes a three-tier classification system (see Table~\ref{tab:mof_classes_components}) to classify the degree of completeness and openness of ML models across all aspects of a model's development lifecycle. The MOF has 17 components to fulfil completeness of model artifacts, which cover the code, data, and documentation that are part of the model development lifecycle (see definitions of each components in Section~\ref{sec:mofcomponents}). %Four components---datasets, supporting libraries and tools, model metadata, and sample model outputs---are optional, with the caveat that datasets must be included for Class I (any or no license). 
The distribution includes an additional component, the MOF configuration file, to comply with the MOF requirements.

The 17 components are categorized into three distinct classes. Each class builds upon the previous one, with Class III being the least complete and and Class I being the most complete (see Table~\ref{tab:mof_classes_components}). The higher the class indicates a more complete distribution that promotes more transparency and enables reproducibility, auditing, and downstream use. This approach is more meaningful than a calculated index, as it guides model producers in providing essential components released under open licenses for each tier of the framework. As the class of the MOF increases, the producer moves closer to a more complete distribution that best aligns with the principles of open science in AI. To qualify for a particular class, the producer must provide every required component for that class. Each component must be released using an appropriate open license from Table~\ref{tab:mof_components_licenses} to qualify the entire project at the specified class level.

\begin{table*}[h]
\centering
\renewcommand{\arraystretch}{1.15}
\begin{tabular}{|p{0.25\linewidth}|p{0.69\linewidth}|}
\hline
\textbf{MOF Class} & \textbf{Components Included} \\ \hline
\textbf{Class III. Open Model} & \makecell[l]{\\
1. Model Architecture\\
2. Model Parameters (Final Checkpoints)\\
3. Technical Report or Research Paper\\
4. Evaluation Results\\
5. Model Card\\
6. Data Card\\
7. Sample Model Outputs (Optional) \\\\
} \\ \hline
\textbf{Class II. Open Tooling} & \makecell[l]{\\
1. All Class III Components\\
2. Training, Validation, and Testing Code \\
3. Inference Code\\
4. Evaluation Code\\
5. Evaluation Data\\
6. Supporting Libraries \& Tools \\\\
} \\ \hline
\textbf{Class I. Open Science} & \makecell[l]{\\
1. All Class II Components \\
2. Research Paper \\
3. Datasets \\
4. Data Preprocessing Code\\
5. Model Parameters (Intermediate Checkpoints)\\
6. Model Metadata (Optional)\\ \\
} \\ \hline
\end{tabular}
\caption{Model Openness Framework Classes and Components}
\label{tab:mof_classes_components}
\end{table*}

\subsection{MOF Class Descriptions}
The 3 classes of the MOF represent ascending levels of model completeness and openness. We describe the distinguishing aspects of each tier beginning with the lowest class.

\subsubsection{Class III. Open Model}
In the MOF, Class III is the entry point and contains the minimum required components that must be released using open licenses. If not all of these components are included in a release and all components do not use an open license then the entire release cannot be considered open under the MOF. The Open Model class covers the following: 
\begin{itemize}
    \item Core model architecture and the final set of parameters
    \item Light documentation conveying capabilities and characterization of the model and data.
\end{itemize}

Class III contains all the components required to study, modify, redistribute, and build upon a model without restrictions, including for commercial and educational purposes. The inclusion of the model architecture, final weights and biases, and documentation (including the technical report, evaluation results, model and data cards) provides the necessary information to work with the model and understand its capabilities, constraints, and the nature of the training data. However, this class lacks completeness and robustness for full reproducibility and the transparency needed to confirm all claims made by the producer. It also lacks sufficient components to evaluate the model, including the training data.

\subsubsection{Class II. Open Tooling}
Building upon Class III, Class II provides model consumers the complete codebase including libraries and tools needed for training, assessing and testing models themselves. Added elements include:
\begin{itemize}
    \item Full training and inference code
    \item Benchmark tests to validate and quantify performance
    \item Libraries and tools to ease integration and to complete the codebase (optional)
\end{itemize}

This tier is an intermediate step between an open model and open science, providing a model consumer with information to test a model producer's assertions. It also allows a model consumer to perform debugging, and it allows for enhancements to model functionality. Although it does provide insights into the training process, it does not include the actual datasets. It is also lighter on documentation, which limits a deeper understanding of the model’s intricacies.

\subsubsection{Class I. Open Science}
The top tier aligns with ideals of open science: the sharing of all artifacts needed for end-to-end transparency, reproducibility, and collaboration. This includes:
\begin{itemize}
    \item A detailed research paper conveying the genesis of the model and its evolution
    \item Raw training datasets used in the training of the model (any license or unlicensed)
    \item Checkpoint weights showcasing full model evolution
    \item Log files providing yet more low-level insights
\end{itemize}

Fulfilling Class I empowers the community to inspect models through the model lifecycle along multiple fronts, representing the gold standard for completeness and openness rooted in scientific principles.

\subsection{Hybrid Releases}
Openness has always been a binary decision in the open-source movement; software is either open-source or not, with no in-between \cite{OSI2007}. A developer either released their software under an OSI-approved license or they did not. If any essential component was not released under an open-source license, the entire release was no longer considered open source. The MOF follows this principle. When any component is not released using an open license as described in Table~\ref{tab:mof_classes_components}, that component is not deemed open and does not qualify for an MOF class. Removing a component that moves the project into a lesser class is acceptable if all remaining components are released with open licenses. 

To qualify as a Class III project, the model, its parameters, and a technical report that describes the work along with evaluation results and model and data cards must be released with open licenses. If not, the project cannot be considered open. This includes projects that use modified open licenses and implement restrictions or acceptable uses.

It should be noted that the MOF classifies models and their components on completeness when they are open. The reader should not confuse the classification system as being a gradient measure of openness \cite{solaiman2023gradient}, but rather a measurement of the completeness of a release in adherence with open science principles \cite{phang2022eleutherai, chesbrough2015open,willinsky2005unacknowledged}.

\section{MOF Components} \label{sec:mofcomponents}
The following defines the 17 components included in the above three-tier classification system of models (see Table~\ref{tab:mof_classes_components}). They cover the degree of completeness and openness across all aspects of the development process of an ML model, including training data, model architecture, model parameters, evaluation benchmarks, and documentation.

Note that not all components are required for all classes. Each component section below specifies the classes and it applies to and Table~\ref{tab:mof_classes_components} lists the components required for each class.

Note that not all components need to be distributed separately, some MAY be combined. For example, evaluation results MAY be included in the research paper, technical report, or model card rather than published as a standalone artifact. This sort of combination SHOULD however be limited to combining component types that are covered by the LICENSE file for that component or the whole distribution. 

The key words "MUST", "MUST NOT", "REQUIRED", "SHALL", "SHALL NOT", "SHOULD", "SHOULD NOT", "RECOMMENDED", "MAY", and "OPTIONAL" in this section are to be interpreted as described in RFC2119 \cite{bradner1997key}.

\subsection{Model Architecture (III.1)}
The model architecture is the core of any ML project. It can include the ML algorithms, neural network layout, connectivity, activations, and other architectural elements. Examples include transformers (e.g., GPT, BERT), convolutional neural networks (CNNs), recurrent neural networks (RNNs), and graph neural networks (GNNs). While the model architecture is often closely tied to the trained model parameters, sharing the architecture alone allows others to understand the structure of the model without necessitating the release of the fully trained model. The model architecture SHOULD be fully described in the research paper, technical report, or model card, and MUST be distributed as open source code under an OSI-approved open source software license that does not limit its usage and derivative works. 

\subsection{Model Parameters – Final Checkpoints (III.2)}
Trained model parameters MUST be released under an open license. In the case of deep learning models, checkpoints from key intermediate stages of training as well as the final optimizer state SHOULD be included. At a minimum the final model parameters and optimizer state (when applicable) MUST be distributed, whether compressed or uncompressed, in a format compatible with popular deep learning frameworks such as TensorFlow, Keras, PyTorch or the framework independent ONNX file format. 

To date, model producers have been releasing model parameters (i.e., weights and biases) using an open source software license, such as Apache 2.0 and MIT, even though model parameters are not compatible with such licenses. Since model parameters are in fact data, model parameters SHOULD be distributed under an open data license, like CDLA-Permissive-2.0. Although licenses designed for OSS are permissive and indemnify the developer from liability, open data licenses are better suited to data-specific considerations such as privacy, ethics, and data rights. Most permissive licenses do not refer to data directly and do not address the ability to modify and redistribute model parameters. This gap could result in a legal obligation to any model consumer if the model producer were to implement royalties after widespread adoption of their model. This is a legal gray area that remains untested. The model architecture and model parameters SHOULD be saved independently in different files for distribution, as each one requires a different format-appropriate open license. This separation allows each component to be studied, modified, redistributed, and used independently of the other.

\subsection{Technical Report (III.3)}
The technical report, which MAY be in the form of a white paper, provides the necessary documentation for the model consumer to understand the performance, usage, and implications of using the model, but not necessarily enough details to reproduce the model and replicate its results. The technical report MUST be included in the distribution or made available on a permanent open access platform such as arXiv, and SHOULD be linked from the model card. The technical report MAY be omitted if a research paper is provided. The technical report MUST be distributed under an open license that SHOULD be appropriate for documentation, ideally CC-BY-4.0 or CC0.

\subsection{Evaluation Results (III.4)}
Detailed quantitative metrics and qualitative results from evaluating the model MUST be reported. They MAY be included in the technical report, the research paper, or the model card. Tests can evaluate any factor, not limited to model efficiency, accuracy, performance, fairness and bias evaluations, toxicity, truthfulness and so forth. Producers MUST include benchmark test results, whether industry standard benchmarks or custom benchmark tests that were developed. If industry standard benchmark tests or test suites are used, the test suite name, test name and version number MUST be included with the results. If custom benchmarks were developed, whether in code or any form of media including text, images, the custom benchmarks MUST be included in full for validation. The raw outputs of the model evaluation MUST be distributed under an open license that SHOULD be appropriate for content like CC-BY-4.0.

\subsection{Model Card (III.5)}
A model card provides metrics, usage guidance, and details about a model \cite{mitchell_modelcards_2019}. Model cards SHOULD cover model details, intended uses, factors, evaluation, risks, and mitigations related to the model. The model card itself MUST be distributed under an open license that SHOULD be appropriate for documentation, ideally CC-BY-4.0.

\subsection{Data Card (III.6)}
A data card provides summary statistics and key information about a dataset to enhance understanding of its composition \cite{gebru2021datasheets}. Following guidelines from the Data Nutrition Project \cite{Data_Nutrition_Project_2021}, data cards SHOULD describe metrics about the features, instances, intended uses, motivation, and collection process. Data cards help identify potential biases in datasets and guide proper usage by downstream usage. They also contribute to reproducibility and transparency by detailing the entire data preparation process. The data card MUST be distributed under an open license that SHOULD be appropriate for documentation, ideally CC-BY-4.0.

\subsection{Sample Model Outputs (III.7)} 
Sample outputs are text samples, images, videos, software code, audio, 3D assets, metadata or any other potential output generated from the model, including predictions and probabilities. Sample outputs are RECOMMENDED but not required. They MAY be omitted. If they are included in the distribution, they MUST be shared publicly without copyright or restrictions where legally permitted so that they can be redistributed with the release. For certain sensitive domains, generated examples MAY be anonymized or simulated if needed. In the event where model outputs are not copyrightable, the outputs SHOULD be released without a license, and this SHOULD be noted in the LICENSE file.

\subsection{Training, Validation, and Testing Code (II.2)}
The full code for training, validating and testing the model, including model construction, training loop, hyperparameter selection, and checkpointing SHOULD be distributed as open source software. Any fine-tuning code, reinforcement learning code or any other method that otherwise modifies that model parameters or code that implements adapters that ultimately affect the performance of the model MUST be included. Comments explaining the approach SHOULD be included in the code, ideally following PEP 8 style guide for Python code. It is also RECOMMENDED to include with the release the log files generated during training. The training, validation, and testing code itself MUST be released under an OSI-approved open source software license while any logs SHOULD be covered by a permissible open content license like CC-BY-4.0.

\subsection{Inference Code (II.3)}
The availability of inference code facilitates complete replication of the performance of the model, and it informs the model consumer about how to use the model most effectively for their applications. Code for performing inference includes any data preprocessing or postprocessing required during inference and possibly any model optimizations and dependencies like external libraries. It MUST include any code required to fully replicate the benchmark results for the model. The inference code MUST be released under an OSI-approved open source software license.

\subsection{Evaluation Code (II.4)}
Evaluation code, evaluation data, and evaluation results are separate components in the MOF. This is due to the fact that some benchmarks are written in code and other benchmarks only use data; for instance, text used to evaluate an LLM or images used to evaluate a computer vision model. Many benchmark tests are a combination of both code and data used to evaluate a model, which includes the scripts needed to load the data and run benchmark tests. Since code and data require different licenses, they are separate components. Depending on the nature of the model and the methods used to evaluate it, the distribution MAY include one or both of evaluation code and data. Any code used for model evaluation and benchmarking MUST be included and distributed under an OSI-approved open source software license.

\subsection{Evaluation Data (II.5)} 
When the model is evaluated with data (be it any media format including text, images, videos, audio, 3D data, and so forth), that evaluation data MUST be included with the distribution. Where the model producer relies on standard benchmark tests that are widely disseminated, they MAY be omitted from the distribution, but they MUST be described in the technical report, research paper, or model card, along with the version of the test. The evaluation data MUST be released under a data or content appropriate open license like CDLA-Permissive-2.0, CC-BY-4.0 or CC0.

\subsection{Supporting Libraries and Tools (II.6)}
Any supporting code libraries, utilities, or tools developed in the course of the development of the model SHOULD be distributed under an OSI-approved open source software license. This includes data loaders, visualization code, simulation environments, etc. Use of existing and custom open source tools SHOULD also be documented. Any of the following tools and libraries SHOULD also be included:

\begin{itemize}
    \item Software libraries and frameworks used in model development along with version details.
    \item Tokenizers: Code used to tokenize text and any data used to train the tokenizer (if used.)
    \item Hyperparameter search code: Code for automating hyperparameter tuning (if used).
    \item Compute infrastructure code: If specialized compute infrastructure was built to scale training, the setup code could be released. 
    \item Monitoring code: Code for tracking experiments, metrics, artifacts etc. during model development is often useful to open source as well. 
    \item Containerization files: Dockerfiles or other container packaging to distribute the model could be shared. 
    \item Frontend/visualization: Any web/mobile frontends or visualizations built on top of the model outputs could be released as open source. 
    \item Deployment orchestration: Infrastructure-as-Code templates for deploying the model to production. 
    \item Model integration code: Wrapper code/SDKs to integrate the model into downstream applications. 
    \item Interactive demos: Links to hosted interactive demos of the model through Jupyter, Streamlit, etc.
\end{itemize}

Presumably most libraries and tools used already have their own licenses, but if the model producer created their own libraries or tools they MUST include them with the distribution under an OSI-approved open source software license.

\subsection{Research Paper (I.2)}
The research paper detailing the model methodology, results, and analysis MUST be included in the distribution or made available on a permanent open access platform such as arXiv, and SHOULD be linked from the model card. The research paper MUST be released under an open license appropriate for documentation, ideally CC-BY-4.0. The following structure is RECOMMENDED but not required: abstract, introduction, related work, methods, results, discussion, conclusion, references. The research paper MAY be replaced with a detailed technical report providing the same information and distributed under an open license appropriate for documentation, ideally CC-BY-4.0.

\subsection{Datasets (I.3)}
Data is the lifeblood of ML models and is the most often held back element in the release of a model. Datasets include training data which is data used for any form of model training including pre-training, fine-tuning, alignment using reinforcement learning techniques or data used for other methods that otherwise modify the weights of the model. Datasets also include data used for model validation and testing as well as data used with benchmark tests. The datasets component also includes tokenized datasets when present. Data can be any form or combination of media, whether text, code, images, videos, audio, 3D objects, URIs and any other data used for training, validation and testing purposes. Datasets also include any metadata. This includes anything from annotation data like labels, bounding boxes and key points to attribution, bitrates, resolution and other metadata relevant to a dataset used in the model development process. The datasets used to develop the model MUST be provided, in the public domain, as copyrighted data, or under any form of license. They SHOULD be released under an open license, preferably CC-BY-4.0 or CC-0. Any limits on sharing due to privacy or sensitivity SHOULD be documented. Both pre- and post-processed data SHOULD be supplied, however producers MAY provide instead links to any curated raw datasets online if they are accompanied by data preprocessing code.

\subsection{Data Preprocessing Code (I.4)}
The data preprocessing code is all code used for preprocessing, cleaning, and formatting the training, validation, and testing data for a model. It also includes code used to transform fine-tuning data and code that is used for alignment tasks like Reinforcement Learning from Human Feedback (RLHF). Other data preprocessing code such as code for data ingestion when appropriate, feature engineering, data augmentation and tokenization is also included. The data preprocessing code MUST be released using an OSI-approved open source software license.

\subsection{Model Parameters – Intermediate Checkpoints (I.5)}
In addition to the final checkpoints and optimizer states, for Class I models, the checkpoints and optimizer states (when applicable) from key intermediate stages of training along with the log files MUST be included and distributed under an open license. Intermediate model parameters SHOULD be distributed under an open data license, such as CDLA-Permissive-2.0.

\subsection{Model Metadata (I.6)}
There are other forms of metadata that can provide additional context about the model, such as the version of the framework used to create it and custom tags or descriptions provided by the developer including model and data lineage information. There is no particular requirement or profile for this type of metadata and it can reveal anything the developer would like to include with the shipped model. This information can help with model management, especially when working with multiple versions of models or conducting experiments. Often the metadata is exported from or loaded by a metadata store. The model metadata MAY be included in the model card, research paper, or technical report. Any model metadata SHOULD be covered by an open data license such as CDLA-Permissive-2.0.

\subsection{Model Openness Configuration File}
The MOF configuration file MUST be included in any distribution. It describes what model components are included in the release and what license covers each component.  The file itself MUST be distributed under an open license and SHOULD be distributed under the CC-BY-4.0 license. 

\section{Model Openness Framework Acceptable Licenses} \label{sec:moflicenses}
Table~\ref{tab:mof_components_licenses} provides an overview of acceptable licenses for each component. The table categorizes each component into one of three domains: Data, Model, or both. Additionally, the content type of each component is classified as data, code, or documentation. The table specifies standard open licenses that should be used for releasing each component, while allowing some flexibility for equivalent licenses. By providing a comprehensive scope, the MOF encourages opening the entire pipeline that produces, evaluates, and applies a model. This approach offers multiple perspectives into the model's inner workings, promoting transparency and reproducibility in open model development.

\begin{table*}[t]
\centering
\small
\setlength\tabcolsep{6pt}
\renewcommand{\arraystretch}{1.15}
\begin{tabular}{|p{0.2185\linewidth}|p{0.11\linewidth}|p{0.117\linewidth}|p{0.454\linewidth}|}
\hline
\textbf{Component} & \textbf{Domain} & \textbf{Content Type} & \textbf{Accepted Open License} \\ \hline
Datasets & Data & Data & Preferred: CDLA-Permissive-2.0, CC-BY-4.0\newline Acceptable: Any including unlicensed \\ \hline
Data Preprocessing Code & Data & Code & Acceptable: OSI-approved \\ \hline
Model Architecture & Model & Code & Acceptable: OSI-approved \\ \hline
Model Parameters & Model & Data & Preferred: CDLA-Permissive-2.0\newline Acceptable: OSI-Approved, Permissive Open Data Licenses \\ \hline
Model Metadata & Model & Data & Preferred: CDLA-Permissive-2.0\newline Acceptable: CC-BY-4.0, Permissive Open Data Licenses \\ \hline
Training Code & Model & Code & Acceptable: OSI-approved \\ \hline
Inference Code & Model & Code & Acceptable: OSI-approved \\ \hline
Evaluation Code & Model & Code & Acceptable: OSI-approved \\ \hline
Evaluation Data & Model & Data & Preferred: CDLA-Permissive-2.0\newline Acceptable: CC-BY-4.0, Permissive Open Data Licenses \\ \hline
Evaluation Results & Model & Documentation & Preferred: CC-BY-4.0\newline Acceptable: Permissive Open Content Licenses \\ \hline
Supporting Libraries \& Tools & Model & Code & Acceptable: OSI-approved \\ \hline
Model Card & Model & Documentation & Preferred: CC-BY-4.0\newline Acceptable: Permissive Open Content Licenses \\ \hline
Data Card & Data & Documentation & Preferred: CC-BY-4.0\newline Acceptable: Permissive Open Content Licenses \\ \hline
Technical Report & Model \& Data & Documentation & Preferred: CC-BY-4.0\newline Acceptable: Permissive Open Content Licenses \\ \hline
Research Paper & Model \& Data & Documentation & Preferred: CC-BY-4.0\newline Acceptable: Permissive Open Content Licenses \\ \hline
Sample Model Outputs & Model & Data or Code & Unlicensed \\ \hline
\end{tabular}
\caption{Model Openness Framework Components and Licenses}
\label{tab:mof_components_licenses}
\end{table*}

\section{Adopting the Model Openness Framework} \label{sec:mofimplementation}
\subsection{MOF Process Overview}
Unlike other frameworks that attempt to dictate how model producers should build and train their models or create a release path on how models should be released, we take a more objective approach by evaluating models based on their completeness and openness. This approach does not constrain model producers into a single methodology but rather lays out a pliable process that acts as a guideline to help model producers create the most complete and open models. At the completion of the process the MOF provides model producers with a badge for their MOF class that clearly demonstrates to the public their commitment to both completeness and openness.

\vspace{2em}

The MOF process generally follows these steps:

\begin{enumerate}
\item Inventory of artifacts
  \begin{enumerate}
  \item Comprehensively list all artifacts involved in creating the model (data, code, documentation, etc).
  \item Capture details like component names, component locations, versions and licenses.
  \end{enumerate}
\item Map to MOF components
  \begin{enumerate}
  \item Align inventory items to the 16 components defined in Section 5.
  \item Multiple inventory elements may map to a single standard component.
  \end{enumerate}
\item Verify licenses
  \begin{enumerate}
  \item For each MOF component present, check if it uses an acceptable open license from Table~\ref{tab:mof_components_licenses}.
  \item If licenses are incompatible, the model cannot be classified.
  \end{enumerate}
\item Determine completeness
  \begin{enumerate}
  \item Check inventory against the component list for the 3 classes in Table~\ref{tab:mof_classes_components}.
  \item Classify model at the highest tier where all required components in the class employ open licenses.
  \item Model meets Class III at a minimum when using open licenses.
  \end{enumerate}
\item Generate MOF.JSON
  \begin{enumerate}
  \item Create the MOF.JSON file, either using the Model Openness Tool (MOT) or manual means.
  \item Include all artifacts, licenses, locations and other required data to meet the MOF requirements.
  \end{enumerate}
\item Self-assert classification
  \begin{enumerate}
  \item With inventory, mapping, and MOF.JSON file finalized, the model producer asserts the appropriate class using the Model Openness Tool (MOT) or through self-assessment.
  \item The model producer must stand behind their completeness and openness claims.
  \end{enumerate}
\item Badging and validation
  \begin{enumerate}
  \item The model producer uses the MOT for badging classified models.
  \item MOT provides the MOF.JSON file and badge code for inclusion with project files.
  \item Community helps ensure accurate labeling by filing disputes.
  \end{enumerate}
\end{enumerate}

This process determines a model's location on the spectrum, guiding model producers in improving openness and consumers in evaluating fitness of models for their usage.

\subsection{Preparing the Distribution}
All projects must include a LICENSE file that describes the licenses used for the project. Conventionally a LICENSE file would include a single license, however it is recommended that the LICENSE file include all licenses that apply to the project. For instance if software is covered under Apache 2.0 and all documentation and data use CC-BY-4.0, then the text of both licenses should be included in the LICENSE file in their entirety including the license heading in order to distinguish what text belongs to which license. Alternatively, a distribution can contain different LICENSE files that are bound to the different components included in the distribution.  Ideally the LICENSE files for each component should be located in the base directory of the component that they cover.  The MOF.JSON file records the path to the appropriate LICENSE file for each component included in the distribution and facilitates both the per component LICENSE method and the single LICENSE file method. 

In addition to the LICENSE file, the distribution must include an MOF.JSON file providing details about the MOF version, release details, included components, and their licenses. This file can be generated with the MOT maintained by the Generative AI Commons at \cite{GenAI_Commons_2024} or created manually or automatically. It is important to note that when a component is not released with the distribution, it should not appear in the MOF.JSON file.  When a component is released but does not use an open license or it uses a custom license, it should not be included in the MOF.JSON file either. The MOF.JSON file only references components that are released using an open license.

\subsection{MOF.JSON Structure}
The MOF JSON file is structured as a single MOF object defined at the root of the JSON file (see GitHub \cite{MOFgithub2023}). Specifically, under the root there are three required, nested objects with their own set of variables:

\begin{itemize}
    \item \textbf{Framework}: This object contains the details related to the framework itself, including the following required variables:
        \begin{itemize}
        \item \textbf{name}: The name of the framework. The variable type is string.
        \item \textbf{version}: The version number of the framework. The variable type is string.
        \item \textbf{date}: The publication date of the framework. The variable type is string in YYYY-MM-DD format.
        \end{itemize}
    \item \textbf{Release}: This object contains the details of the model being released. There are a number of variables:
        \begin{itemize}
        \item \textbf{name}: The name of the release. The variable type is string.
        \item \textbf{version}: The version of the release, which can be the parameter count or another identifier that distinguishes the model from previous versions and versions of the same model with different parameter counts. The variable type is string.
        \item \textbf{date}: The date of the release. The variable type is string in format ``YYYY-MM-DD``.
        \item \textbf{type}: The nature of the model, i.e., language model, image generation, audio generation, image classification, statistical ML, or any number of other types of models. The variable type is string.
        \item \textbf{architecture}: The model architecture employed, i.e., transformer, diffusion, GAN, NERF, VGG, Resnet, K-means, or any other type of model architecture. The variable type is string.
        \item \textbf{treatment}: Any type of post-training treatment, like fine-tuning, constitutional alignment, RLHF or any other treatment that otherwise modifies the parameters of the original model. If no treatment has been applied then this variable is an empty string. The variable type is string.
        \item \textbf{origin}: The original model, generally this is the foundation model. If this is not a foundation model in the release, then this variable contains the name and version of the model that was modified. The variable type is string or left empty for foundation or non-derivative models.
        \item \textbf{producer}: The name model producer or publisher, could be a company, organization, group or individual. The variable type is string.
        \item \textbf{contact}: The email address for the model producer or publisher. The variable type is string.
        \item \textbf{mof\_class}: The qualifying MOF class of the release as generated by the Model Openness Checker. The variable type is integer.
        \end{itemize}
    \item \textbf{Components}: This object contains a list of components that are included with the model distribution, as well as each component's details:
        \begin{itemize}
        \item \textbf{description}: A text description of the component. Using the default values is acceptable. When introducing a new component beyond the standard components, include a description of the component. 
        \item \textbf{location}: The location of the component within the distribution, full path is required in UNIX format with leading slash for the root directory. The variable type is string.
        \item \textbf{license}: The SPDX identifier of the license(s) used for the component. If multiple licenses are used for a single component, often the case for libraries and tools, they must be provided in a comma-separated list. The value must use a valid SPDX license identifier \cite{SPDX_LF}. The variable type is string.
        \item \textbf{license\_path}: The location of the LICENSE file for the component within the distribution, full path is required in POSIX format with leading slash for the root directory. More than one component can point to the same LICENSE file. In the event the component employs multiple licenses, the LICENSE file should contain the text for all the licenses used.  Alternatively, multiple license files may be specified, each separated by a comma. However they must correspond in order to the comma separated list of license names provided in the license variable. The variable type is string.
        \end{itemize}
\end{itemize}

\subsection{Class Assignment}
The MOF relies on self-reporting and projects are not classified by a central authority. LF AI \& Data Generative AI Commons will provide a web interface, the MOT, that allows model producers to fill out a web form with the details of their project and in turn the MOT informs the user how their project lines up with the classes in the MOF. 

\subsection{Model Openness Tool}
The Model Openness Tool (MOT) has been developed to complement the Model Openness Framework (MOF), providing a reference implementation designed to evaluate ML models against the principles outlined by the MOF and to ensure clarity on the permissible uses and restrictions of the model and its various parts. Accessible via \url{https://isitopen.ai/}, the MOT provides three core functionalities that enable users to (1) comprehend the completeness and openness of ML models registered in the MOT catalogue, (2) evaluate the openness of their own models based on released components and associated licenses, and (3) submit models to the MOT catalogue. For registered, evaluated, and submitted models alike, the MOT assesses each criterion from the MOF and generates an openness score based on the degree to which each criterion is fulfilled. By offering a practical and user-friendly mechanism, the MOT facilitates the application of the MOF to any ML model, producing a clear, self-service score. 

\subsubsection{View Models}
\begin{figure}[h]
  \centering
  \includegraphics[width=0.99\textwidth,height=0.55\textheight]{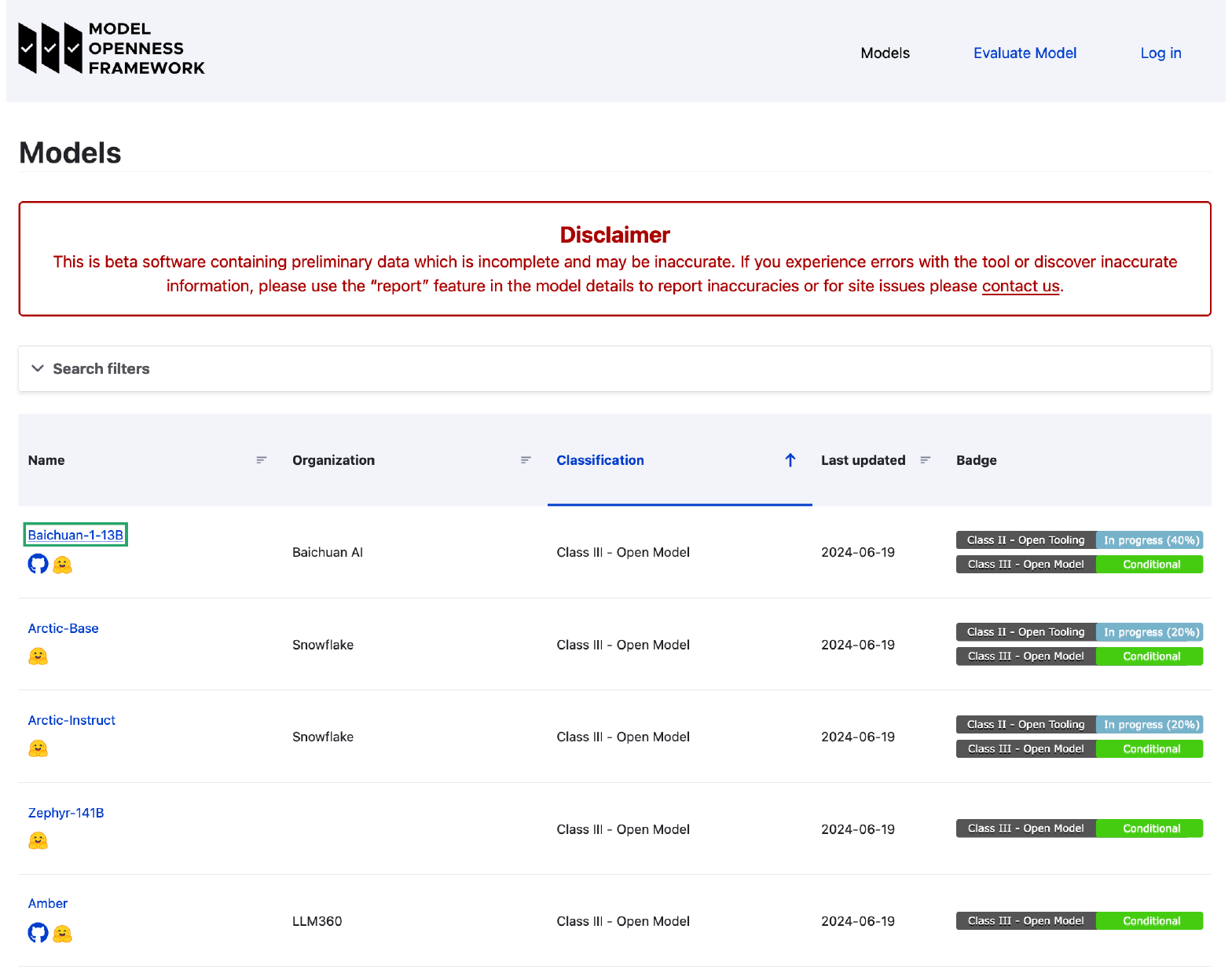}
  \caption{View models in the catalogue of the Model Openness Tool}
  \label{fig:MOT-Catalogue}
\end{figure}

\begin{figure}[h]
  \centering
  \includegraphics[width=1.05\textwidth,height=0.55\textheight]{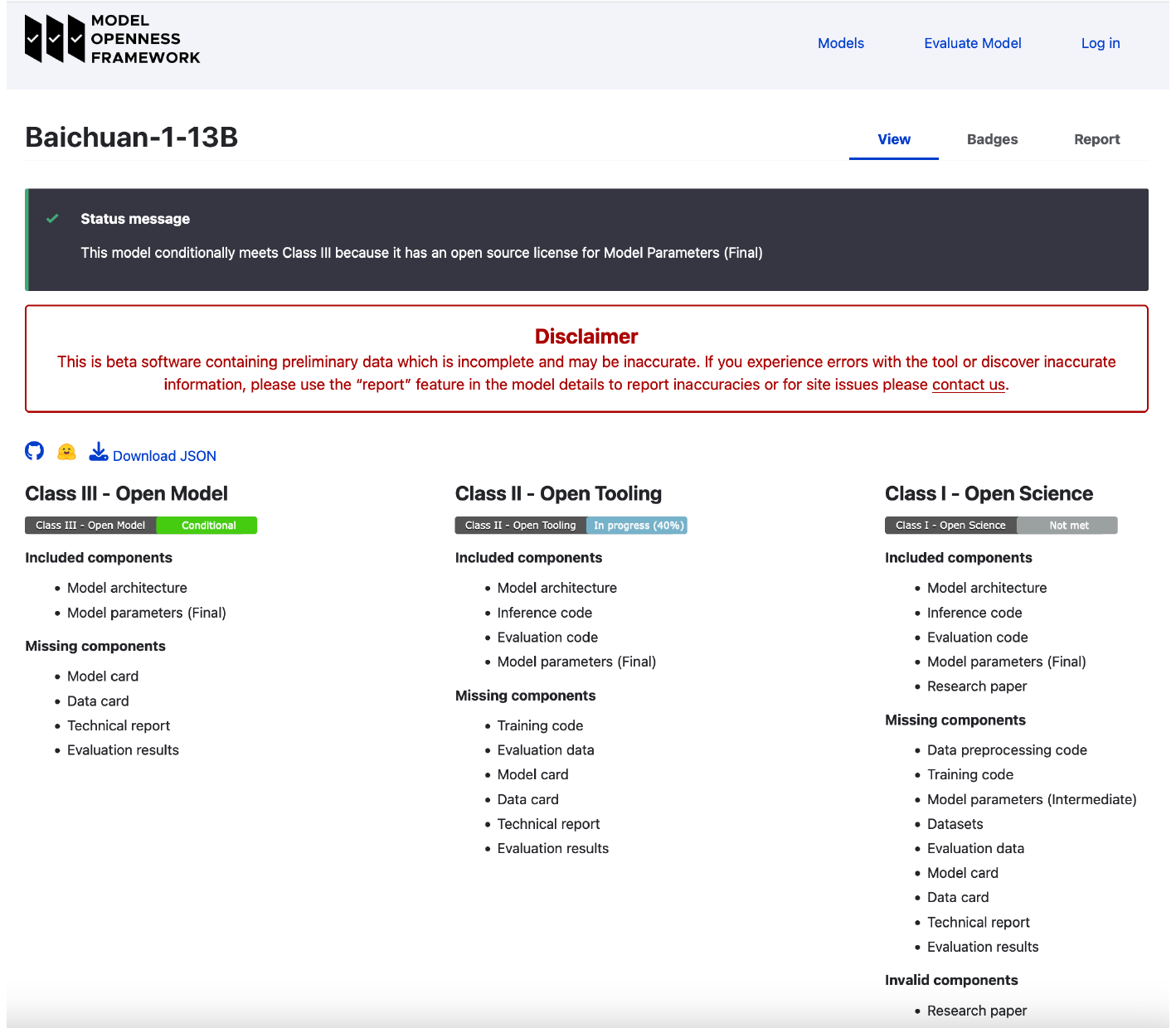}
  \caption{View a model's classification with the Model Openness Tool}
  \label{fig:MOT-ViewModel}
\end{figure}

The MOT catalogue interface (see Figure~\ref{fig:MOT-Catalogue}) presents a tabular view of registered ML models. Each row represents a distinct model, with columns providing key information at a glance. The model name column includes clickable icons that link to the model's repositories on GitHub and Hugging Face Hub, facilitating immediate access to the model's source. The classification and badge are dynamically generated based on the released components and their associated licenses.
Upon selecting a model, users are directed to a detailed model page (see Figure~\ref{fig:MOT-ViewModel}), which provides:

\begin{itemize}
    \item A comprehensive overview of the model's MOF classification.
    \item A component-wise breakdown, categorizing each into: 
    \begin{itemize}
    \item Released with valid licenses,
    \item Released with invalid licenses, or
    \item Unreleased
    \end{itemize}
    \item A copyable MOF badge for external use (e.g., in repositories).
    \item A reporting mechanism for data corrections or updates.
\end{itemize}

\subsubsection{Evaluate Models}
\begin{figure}[h]
  \centering
  \includegraphics[width=0.99\textwidth,height=0.53\textheight]{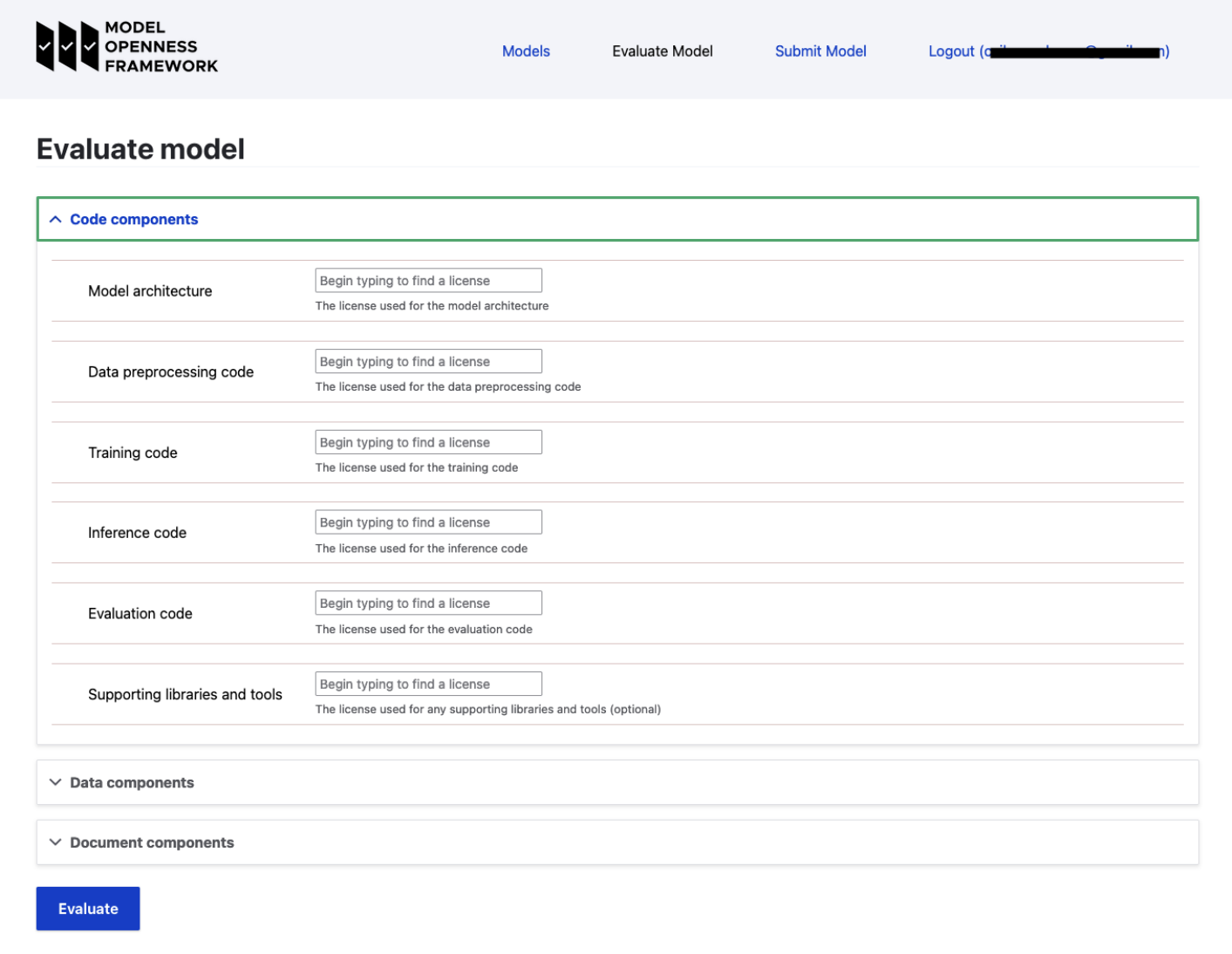}
  \caption{Evaluate models with the Model Openness Tool}
  \label{fig:MOT-Evaluate}
\end{figure}

The evaluation interface (see Figure~\ref{fig:MOT-Evaluate}) allows users to assess the completeness and openness both self-developed and unregistered models. The process involves:
\begin{itemize}
\item Input of license information for each of the 16 MOF components via a dropdown menu.
\item Automatic classification of components with empty license fields as unreleased.
\end{itemize}

Post-evaluation, the MOT generates:
\begin{itemize}
\item A model page with a MOF classification score (1-3).
\item A component-wise breakdown (as in the catalogue model pages).
\end{itemize}

This score provides a quantitative measure that facilitates easy interpretation of a model's alignment with the principles of openness and objective comparisons between models.
% Once the user has inserted relevant information, the tool calculates an openness score that classifies the model’s openness on a scale of 1, 2, or 3, providing an easily ,

\subsubsection{Submit Model}
\begin{figure}[h]
  \centering
  \includegraphics[width=0.99\textwidth,height=0.74\textheight]{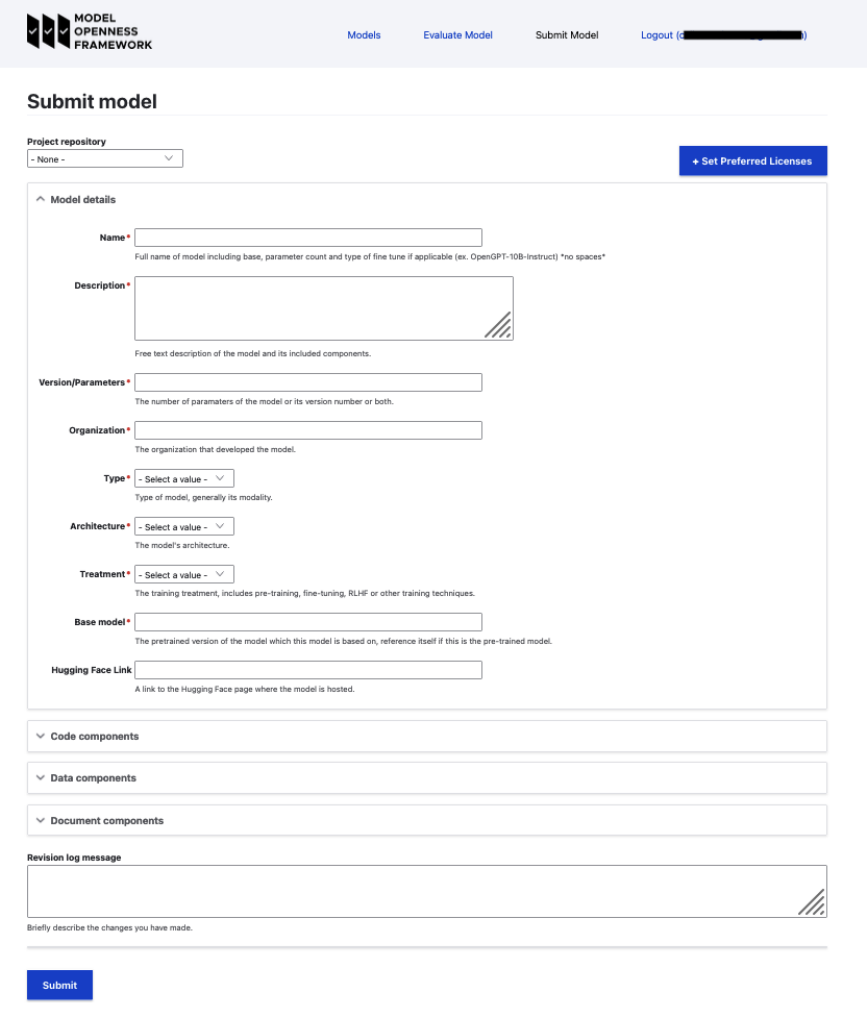}
  \caption{Submit models to the model catalogue with the Model Openness Tool}
  \label{fig:MOT-Submit}
\end{figure}

The submission interface (see Figure~\ref{fig:MOT-Submit}) guides users through a structured process to add models to the MOT catalogue. Key steps include:
\begin{itemize}
\item Input of model metadata, including:
\begin{itemize}
\item Name
\item Description
\item Version/parameters
\item Organization
\item Type (e.g., language model, image model, code model, etc.)
\item Version/parameter count
\item Architecture (e.g., transformer, diffusion, RNN, CNN, etc.)
\item Treatment (e.g., pre-trained, instruct fine-tuned, or chat fine-tuned)
\item Base model
\item Hugging Face Hub link (if applicable)
\end{itemize}
\item License specification for each of the 16 MOF components via dropdown menus.
\end{itemize}

Upon submission, the MOT:
\begin{itemize}
\item Calculates the MOF classification score (1-3).
\item Generates a model page.
\item Integrates the model into the public MOT catalogue.
\end{itemize}

This streamlined process ensures consistency in model representation and facilitates the expansion of the MOT database.

\subsection{Badging System}
The MOF is designed to be both informational and actionable. As such the Generative AI Commons is implementing a badging program, similar to the OpenSSF Best Practices Badge Program \cite{OpenSSF_2021}. The badging system is a part of the MOT \cite{GenAI_Commons_2024}, and is a free service that allows model producers to perform the following:
\begin{itemize}
    \item Perform a check the completeness and openness of their model distribution and display which MOF class their model meets
    \item Receive recommendations on which licenses to use for which components
    \item Generate an MOF.JSON file for their distribution
    \item Be provided with code to insert into their README.md file in their Github repository
    \item Track their model’s ranking amongst other models on the MOF scoreboard
\end{itemize}

For model consumers, they can do the following:

\begin{itemize}
    \item View the MOF scoreboard to see which models are the most complete and open
    \item Drill down into model distributions to see which ones meet their completeness and openness requirements 
    \item Quickly see which MOF class a model has attained in the project’s Github repo
    \item Validate that a model has attained an MOF class
    \item Submit a dispute if they believe that a model is being misrepresented as complete or open
\end{itemize}

It is incumbent upon the producer of an ML model and its components to accurately include the results of either the MOT or accurately identify the components and licenses included in the distribution in the MOF.JSON file and specify the class the project qualifies for. Misrepresentations will only harm the reputation of the model producer.

\subsection{Disputes}
The MOF relies on the honesty and transparency of researchers and developers to accurately classify models and to state which components with which licenses they include.  Therefore, we also rely on the community to identify projects that have been misrepresented as open and notify the organization that hosts the project about their concerns. 

\section{Benefits of the MOF} \label{sec:mofbenefits}

The adoption of the MOF by the AI community brings many advantages, including but not limited to: 

\begin{itemize}
    \item \textbf{Clarity:} Clearly defines what components are included and under which license each is distributed, in order to understand the acceptable forms of use and whether a project is complete and truly open or not. 
    \item \textbf{Openness:} By classifying models and their artifacts at increasing degrees of openness, the MOF will help push model producers towards creating the most complete and open models, helping to advance open science and both academic and commercial usage.
    \item \textbf{Reproducibility:} Comprehensive availability of data, code, and models enables others to independently reproduce results and identify sources of errors, bias or disparities. This strengthens scientific rigor.
    \item \textbf{Transparency \& Explainability:} Opening model architectures, weights, training code, and documentation sheds light on how models work and behave. This builds appropriate trust and aids inspecting for issues.
    \item \textbf{Data Provenance:} Origination and attribution can be determined when the data and its details are released. This can be helpful in tracing bias in models or identifying sources of PII leakage.
    \item \textbf{Accountability \& Fairness:} Public data and models can be audited for unwanted biases and harms. Model producers can be notified of problems discovered by the community.
    \item \textbf{Continuous Improvement:} Model producers and consumers can build on open models instead of starting from scratch, accelerating innovation and progress in AI.
    \item \textbf{Collaboration:} Sharing open resources allows model producers and consumers across different fields and organizations to pool knowledge and capabilities.
    \item \textbf{Education \& Learning:} Data, code, and models support teaching and learning about AI. Students and new researchers and developers can more easily enter the field.
    \item \textbf{Regulation:} Openness makes models more amenable to oversight and governance, unlocking policy options.
\end{itemize}

\section{Limitations and Criticisms} \label{sec:moflimitations}

\subsection{Known Limitations} 
We acknowledge several limitations and likely criticisms:

\begin{itemize}
    \item The MOF is designed for deep learning artifacts but does not transfer directly to every form of learning in AI. It is applicable to classical ML but does not translate entirely to all aspects of reinforcement learning.
    \item Model producers are expected to be honest about the availability of the components released with their models and the openness of licenses for each component as well as the completeness of both in their release.
    \item It requires convincing model producers who may be reluctant to share their work publicly without restrictions.
    \item Openness goals must be balanced with privacy, IP, institutional policies, and commercialization pressures. 
    \item Classifying models ignores their actual functionality, and bias, safety, and other harms remain a concern. However, openness with models and data enables external audits of quality and completeness.
    \item Simplicity of classification may not capture all nuances. But enhancement of the rubric may occur.
    \item It does not address the use of copyrighted materials in training data, an area currently being addressed through courts and legislation. The MOF requires data to be open using an open license; however, we encourage model producers to use authorized data in training models and respect copyrights \cite{Podnar_2023}.\end{itemize}

\subsection{Out of Scope} \label{sec:outofscope}
The MOF is not designed to solve all issues related to AI and openness, and its effective adoption will rely on the AI community to be transparent and honest in their reporting of the components of the models that they release and the licenses applied to each. The MOF does not intend to address any of the following as they are best addressed through alternative methods and means: AI safety (including bias, fairness, and trustworthiness), performance testing, red-teaming, security and privacy, components related to model serving, and model provenance.

\section{Conclusion}\label{sec:conclusion}
The MOF provides a clear methodology for evaluating and enhancing the openness and completeness of ML models. It outlines specific components that should be openly released, including training data, code, model architecture, model parameters, and documentation, among others, as well as with which licenses. This framework gives model producers a roadmap to follow for reproducible and transparent AI development. By adopting open licenses, as prescribed by the MOF, we can foster collaboration and community engagement, allowing the freedom to use, study, modify, and distribute models and components under the terms of its license. Furthermore, the tiered classification system incentivizes releasing models with increasing levels of completeness. The widespread adoption of the MOF promises to establish completeness and openness as core tenets of responsible AI, ultimately promoting a more transparent and trustworthy advancement of AI R\&D. We encourage model producers to incorporate the framework into their policies to make open science the standard for model distribution, as well as the wider AI community to recognize and reward complete and open distribution of models. Realizing this vision requires a concerted effort by diverse AI stakeholders to embrace both completeness and openness as core tenets of AI R\&D. However, the benefits for science, innovation, and society make pursuing model openness worth the challenge. With carefully designed incentives, policies, and community norms, open source and open science ideals can become the norm in AI R\&D, rather than the exception.

\newpage
%TC:ignore
%Bibliography
\bibliographystyle{ieeetr}  
\bibliography{refs}

\newpage
\appendix
\section{Functionalities of the Model Openness Tool}

\newpage

\newpage

\newpage

\newpage
\section{Acknowledgements}
The following members of the Generative AI Commons were instrumental in reviewing and improving upon the MOF, and their time and contributions are greatly appreciated:  Stella Biderman, EleutherAI; Justin Colannino, Open Source Initiative; Stefano Maffulli, Open Source Initiative; Ke Ding, Intel; Ann Mary Roy, HP; Anni Lai, Futurewei; Ofer Hermoni, Generative AI Commons; Nick Chase, Cloud Geometry; and Saurabh Tangri, Intel.

\section{About the Generative AI Commons}
The Generative AI Commons is a community-driven initiative at the Linux Foundation’s AI \& Data Foundation. It is a vendor neutral forum and open participation initiative focused on advancing principles of open science and open source in generative AI. The Generative AI Commons is dedicated to fostering the democratization, advancement and adoption of efficient, secure, reliable, and ethical Generative AI open source innovations through neutral governance, open and transparent collaboration and education.  More about the Generative AI Commons, as well as details and links to join the community can be found at \url{https://genaicommons.org}.

\section{About the LF AI \& Data Foundation}
The LF AI \& Data Foundation is a global not-for-profit foundation that hosts critical components of the global AI \& data technology infrastructure at the Linux Foundation. It brings together the world’s top developers, end users, and vendors to identify and contribute to the projects and initiatives that address industry challenges for the benefit of all participants. More about the LF AI \& Data Foundation can be found at \url{https://lfaidata.foundation/}
%TC:endignore
\end{document}